\documentclass[conference]{IEEEtran}
\ifCLASSINFOpdf
\else
\fi
\def\begeq{\begin{equation}}
\def\endeq{\end{equation}}
\def\begcs{\begin{cases}}
\def\endcs{\end{cases}}

\newcommand{\nof }[1]{\ensuremath{\mathsf{n}_\text{#1}}}

\def\kmeans{K-Means}
\def\scorMelRGB{$s_{\text{RGB}}$}
\def\lesconf{$L$}
\def\lesconfCRA{$L_{\text{CRA}}$}
\def\lesconfIty{$L_{\text{Ity}}$}
\def\lesconfHull{$L_{\text{Hull}}$}
\newcommand{\hstAtt}{\ensuremath{H^{\text{I}}}}

\usepackage{graphics,epsfig,epstopdf,verbatim,url} % for postscript graphics files
\usepackage{wrapfig,amssymb,amsmath,amsfonts,bm} % ,algorithm,algorithmic}

\begin{document}
% SSSSSSSSSSSSSSSSSSSSSSSSSSSSSSSSSSSSSSSSSSSSSSSSSSSSSSSSSSSSSSSSSSSSSSSSSSSSSSSSSSSSSSSSSSSSSS
%										TITLE & ABSTRACT 										
% SSSSSSSSSSSSSSSSSSSSSSSSSSSSSSSSSSSSSSSSSSSSSSSSSSSSSSSSSSSSSSSSSSSSSSSSSSSSSSSSSSSSSSSSSSSSSS
\title{Melanoma Recognition with an Ensemble of Techniques for Segmentation and a Structural Analysis for Classification}

\author{\IEEEauthorblockN{Christoph Rasche}
\IEEEauthorblockA{Image Processing and Analysis Laboratory\\University "Politehnica" of Bucharest\\Bucharest 061071, Romania. \\
rasche15 at gmail com}
}
%\IEEEoverridecommandlockouts
%\IEEEpubid{\makebox[\columnwidth]{978-1-4673-8200-7/15/\$31.00~
%\copyright2015
%IEEE \hfill} \hspace{\columnsep}\makebox[\columnwidth]{ }}
\maketitle

\begin{abstract}
An approach to lesion recognition is described that for lesion localization uses an ensemble of segmentation techniques and for lesion classification an exhaustive structural analysis. For localization, candidate regions are obtained from global thresholding of the chromatic maps and from applying the \kmeans~algorithm to the RGB image; the candidate regions are then integrated. For classification, a relatively exhaustive structural analysis of contours and regions is carried out. 
\end{abstract}

% \fi
%

\IEEEpeerreviewmaketitle

% SSSSSSSSSSSSSSSSSSSSSSSSSSSSSSSSSSSSSSSSSSSSSSSSSSSSSSSSSSSSSSSSSSSSSSSSSSSSSSSSSSSSSSSSSSSSSS
%										I N T R O D U C T I O N 										
% SSSSSSSSSSSSSSSSSSSSSSSSSSSSSSSSSSSSSSSSSSSSSSSSSSSSSSSSSSSSSSSSSSSSSSSSSSSSSSSSSSSSSSSSSSSSSS
\section{Introduction}

There are two short-comings with lesion recognition systems using conventional techniques (conventional = not Deep Neural Networks): one is that they often attempt to carry out a task with a single technique, in particular for segmentation; another one is, that the structural analysis is often hesitant. For segmentation we show that the combination of multiple techniques produces results much better than any of the individual techniques. For classification we show that the use of a large number of structural parameters has an enormous potential. Both is demonstrated in the melanoma competition, “ISIC 2018: Skin Lesion Analysis Towards Melanoma Detection” using the data as described in \cite{codella2018ISIC,tschandel2018ham}. 

\paragraph{Lesion Segmentation}
Segmentation using conventional techniques used to be carried out with a variety of general segmentation techniques, sometimes using sophisticated methods such as active contours or normalized cuts \cite{korotkov2012computerized,scharcanski2014computer,mishra2016overview}. Recently there has been a shift to combining multiple techniques, because a single technique often fails in certain circumstances, for which another technique can perform well. For instance, Celebi et al. use multiple single (global) threshold methods \cite{celebi2013lesion}; Neghina et al. combine the method of region growing with a method of pixel clustering \cite{neghinua2016automatic}. In this study, we combine multi-level (global) thresholding with a pixel-clustering method (Section \ref{s_Segmentation}).

 %Using such an ensemble of techniques is analogous of using multiple classification algorithms for solving a classification task \cite{Alpaydin12}.

\paragraph{Classification using a Structural Analysis}
Previous conventional approaches to classification had focused on some color and texture features \cite{mishra2016overview,barata2014two} but did not make an effort to extract a wider range of structural descriptors. Here we employ our methodology, that we recently have succesfully applied to satellite images, with accuracies in the range of Deep Neural Networks \cite{Rasche2018satimg}.

% SSSSSSSSSSSSSSSSSSSSSSSSSSSSSSSSSSSSSSSSSSSSSSSSSSSSSSSSSSSSSSSSSSSSSSSSSSSSSSSSSSSSSSSSSSSSSS
%										SEGMENTATION										
% SSSSSSSSSSSSSSSSSSSSSSSSSSSSSSSSSSSSSSSSSSSSSSSSSSSSSSSSSSSSSSSSSSSSSSSSSSSSSSSSSSSSSSSSSSSSSS
\section{Segmentation}\label{s_Segmentation}

% SSSSSSSSSSSSSSSSSSSSSSSSSSSSS 	METHOD 	SSSSSSSSSSSSSSSSSSSSSSSSSSSSSSSSS
\subsection{Method}
\begin{figure}[!htb]
\begin{center}
	\includegraphics[width=9.7cm]{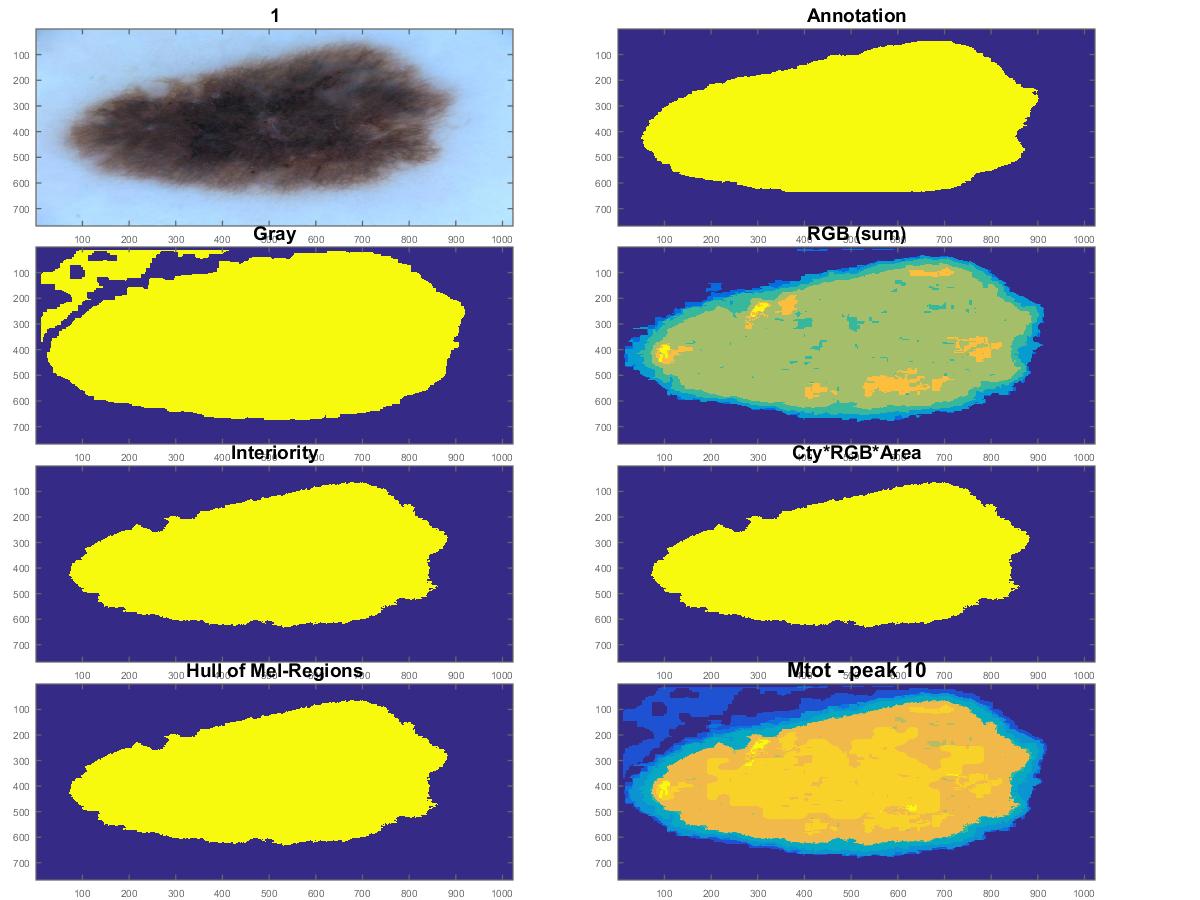}
	\caption{An easy example for segmentation: \textbf{Top Left}: original image. \textbf{Top Right}: provided, manual annotation. \textbf{2nd in Left Column}: Gray-level map: candidate regions as obtained from the analysis of the gray-level histogram (several band thresholds applied). \textbf{2nd in Right Column}: summed output of chromatic-level histogramming (individual output maps are summed). \textbf{3rd Left}: selection of the \kmeans-cluster with highest \lesconfIty~(lower left in Fig. \ref{fig_FlecksEasy}). \textbf{3rd Right}: selection of the \kmeans-cluster with highest \lesconfCRA~(upper right in Fig. \ref{fig_FlecksEasy}). \textbf{Bottom Left}: selection of a \kmeans-cluster with maximal \lesconfHull~(lower right in Fig. \ref{fig_FlecksEasy}). \textbf{Bottom Right}: Integrated `confidence' map: all maps summed to a single map.}
	\label{fig_SegmEasy} 
\end{center}
\end{figure}

Potential lesion regions are detected with two principal methods: with global thresholding of chromatic maps and with clustering of pixels in the three-dimensional RGB space. Fig. \ref{fig_SegmEasy} gives an overview. Several tens of regions are collected per image and to sub-select regions corresponding to actual lesion candidates, we sub-select the regions based on a confidence measure \lesconf. The confidence measure is composed of several parameters. Different confidence measures with varying parameters are used but they essentially all share the parameter centrality $c$ (inverse of the eccentricity from image center) and the parameter area $a$ of the region. Typically the parameters are normalized and multiplied: \lesconf$=c~a$. The lesion candidates of all segmentation techniques are then summed to a single confidence map (lower right in Fig. \ref{fig_SegmEasy}), which then is thresholded using a fixed value, returning a final lesion candidate region (final region not shown in Figure). 

% BF -----------------------------     SEGMENTATION EXAMPLE EASY 
% EF -----------

% ssssssssssss Regions from Thresholding 	ssssssssssss
\subsubsection{Regions from Thresholding}
Multi-level (global) thresholds are applied to the gray-level intensity image and to the individual chromatic-level intensity images. For each such image, the intensity histogram is determined and the major peaks located with their corresponding width. At each peak a band map is generated by thresholding at the two values half the width from the peak value. For each such binary map its regions are determined. To sub-select potential lesion candidates, the confidence measure \lesconf~is applied. 

Fig. \ref{fig_SegmEasy} shows that the process for gray-level segmentation sometimes produces several candidates, see 2nd graph from top in left column. For the output of the chromatic-level segmentations we show only the summed map of all candidates of all three channels, see 2nd graph from top in right column; several smaller regions are detected inside the larger region. Eliminating peripheral or smaller candidates runs the risk of loosing candidates in fragmented lesions, i.e. fragmented by the presence of hair. In the case shown, the presence of the peripheral and smaller candidates had no influence on the final lesion candidate, as the total map shows substantial overlap to segment the actual lesion outline properly.

% Gray-level and chromatic-level segmentation produces several candidates per selection process, at most around ten.

% ssssssssssss Pixel-Clustering 		ssssssssssss
\subsubsection{Regions from Pixel-Clustering}
\begin{figure}[!htb]
\begin{center}
	\includegraphics[width=9.7cm]{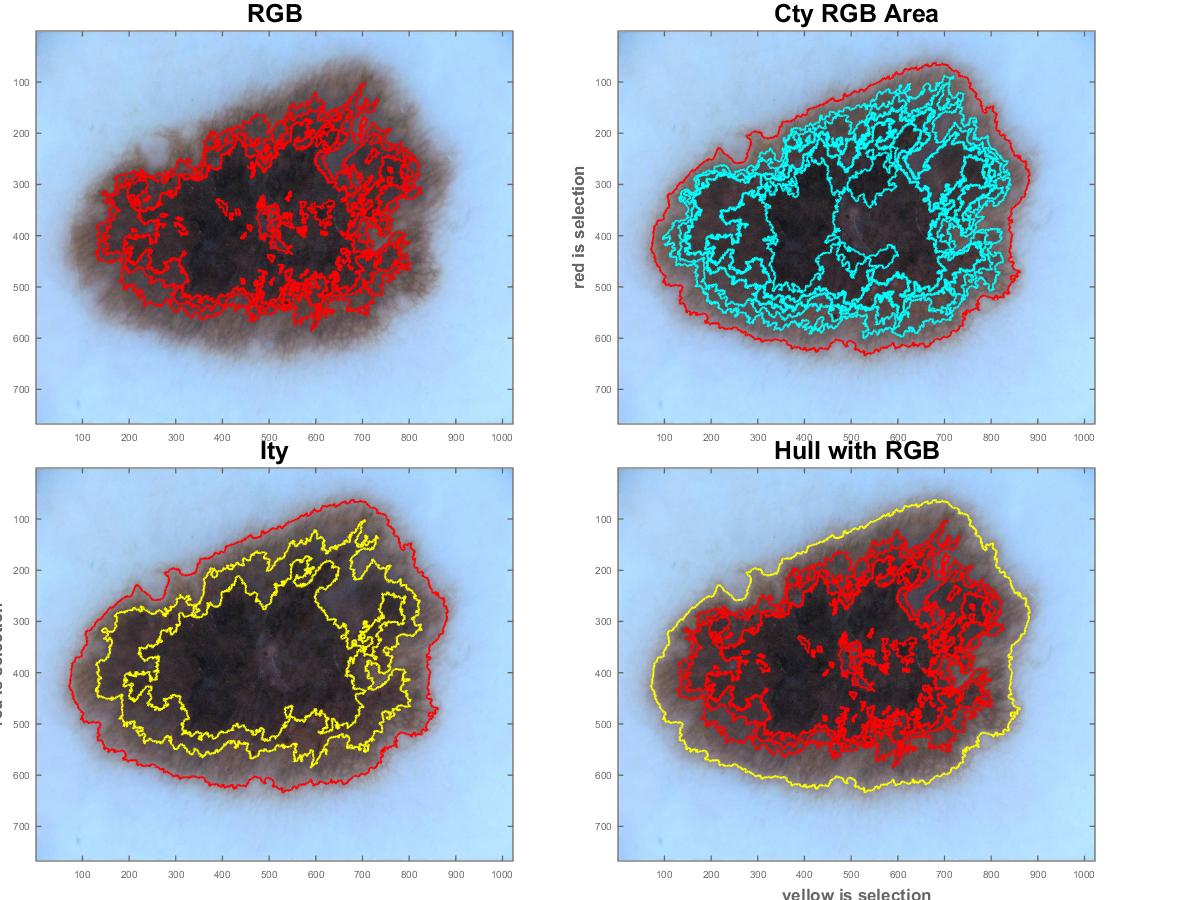}
	\caption{Selections of regions as obtained from the clustering output (\kmeans~on RGB for $k \in [2,3,4,5,6,7,8]$ ). \textbf{Top Left}: regions whose average RGB values correspond to melanoma RGB values (of melanoma regions in other images). \textbf{Top Right}: regions, whose basic lesion confidence \lesconf~is added a parameter expressing the degree of `melanoma color' (resulting in measure \lesconfCRA); the red outline is the region with the highest confidence. \textbf{Bottom Left}: regions that represent the interior \lesconfIty~of the lesion in particular, regions with high solidity and large central compactness. \textbf{Bottom Right}: the region (yellow) with most nested regions that show a high degree of melanoma-like RGB values (measure \lesconfHull).}
	\label{fig_FlecksEasy} 
\end{center}
\end{figure}

The \kmeans~algorithm is employed to cluster pixels in RGB space. The algorithm is applied with a limited range of $k$s, i.e. [2,3,4,5,6,7,8]. For each clustering outcome, its regions are determined; a region consists of a set of contiguous pixels. Regions obtained from pixel-clustering are more diverse in structure than the regions obtained from thresholding: regions from clustering are sometimes fragmented, for instance when the lesion consists of an agglomeration of blobs; or they represent only the lesion silhouette by a ring. For that reason, several types of lesion confidence measures are applied. Those regions are also used to determine the RGB values for regions that lie completely inside the provided, manually annotated region. Those melanoma-RGB values are then matched to region RGB values in other images to obtain a degree of `RGB-melanomaty' for a region, \scorMelRGB. In the top left graph of Fig. \ref{fig_FlecksEasy}, those regions with a high \scorMelRGB~are shown. 

\noindent Three types of lesion confidence measures are developed:

\noindent 1. \lesconfCRA: the measure selects candidates based on the basic lesion confidence multiplied by the degree of RGB-melanomity, \lesconfCRA$ = c~a~$\scorMelRGB. In the top right graph of Fig. \ref{fig_FlecksEasy} some candidates with high \lesconfCRA~are shown, the red one is the one with maximal \lesconfCRA. 

\noindent 2. \lesconfIty: the measure aims at finding the interior of the lesion. This confidence measure was developed to avoid the selection of ring regions, that represent the lesion silhoutte, because those ring regions often contain a substantial part of the skin and we suspected that the interior would be a better lesion representation. The confidence measure \lesconfIty~is based on a number of parameters such as solidity of the region, compactness in the center, etc., all of which are multiplied into the basic lesion confidence \lesconf. Fig \ref{fig_FlecksEasy} lower left shows some candidates with high \lesconfIty, of which red is again the one with maximal value.

\noindent 3. \lesconfHull: the measure prefers regions that contain nested regions with high degree of \scorMelRGB. Fig \ref{fig_FlecksEasy} lower right shows the candidate with maximal value.

% BF -----------------------------     FLECKS EASY 
% EF -----------

% BF -----------------------------     SEGMENTATION EXAMPLE DIFFICULT 
\begin{figure}[!htb]
\begin{center}
	\includegraphics[width=9.7cm]{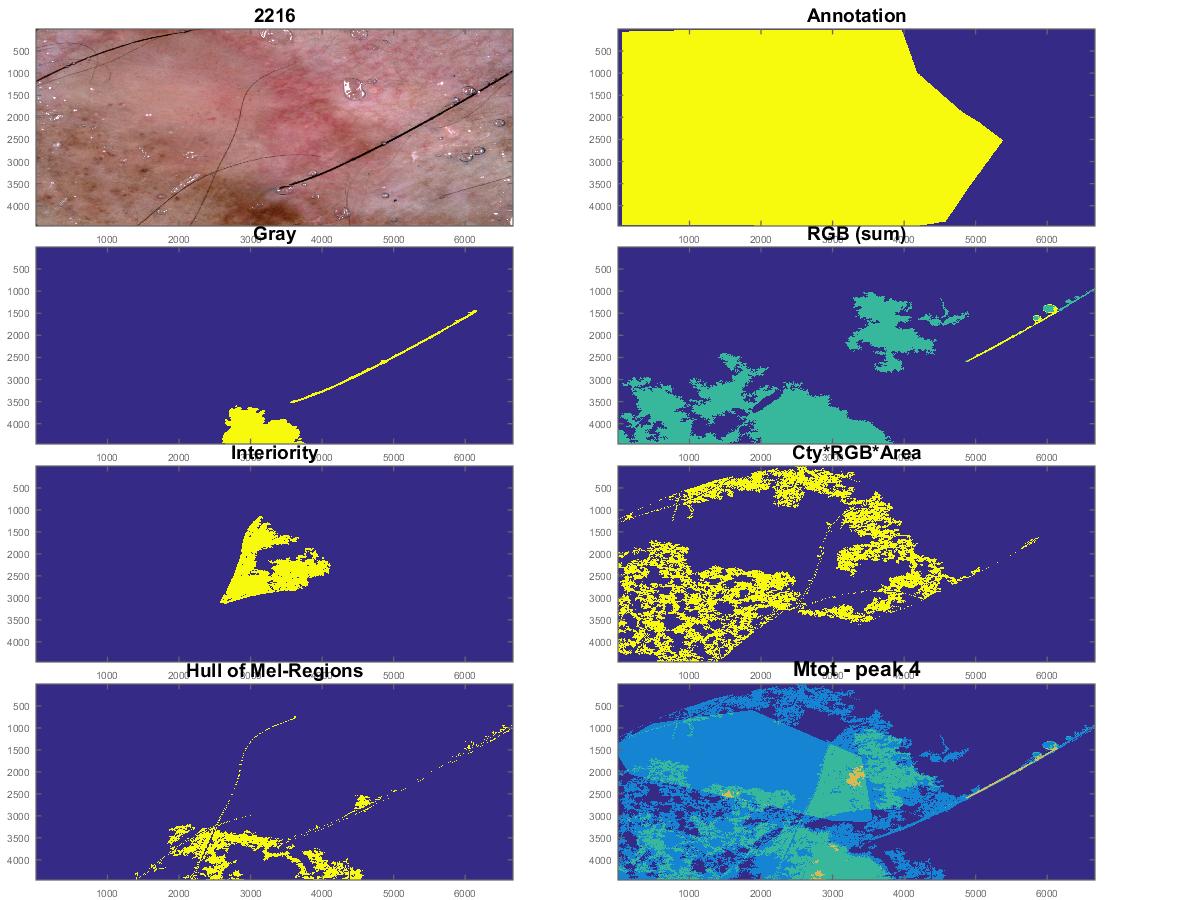}
	\caption{A difficult example for segmentation. Same graph arrangment as in Figure \ref{fig_SegmEasy}.}
	\label{fig_SegmDiff} 
\end{center}
\end{figure}
% EF -----------

% ssssssssssss Ensemble 		ssssssssssss
\subsubsection{Ensemble}
For each of the 7 types (gray, 3 chromatic, 3 cluster regions), a binary map is generated with its most suitable candidates. An individual map often contains one to two candidates, rarely none, sometimes more. Those maps are summed to form the confindence map, see lower right in Figs. \ref{fig_SegmEasy} and \ref{fig_SegmDiff}. That map is then thresholded at some value to arrive at the final candidate region representing the best guess for the lesion location. If there are several candiate regions present in the thresholded map, then the convex hull of all those regions is taken.

% SSSSSSSSSSSSSSSSSSSSSSSSSSSSS 	RESULTS 	SSSSSSSSSSSSSSSSSSSSSSSSSSSSSSSSS
\subsection{Results}
To adjust the parameters, we optimized for accuracy (sensitivity $\times$ specificity) on the set of training images using the provided manual annotation. Adjustment concerned in particular how to deal with candiate regions touching the image borders. The adjustment occured largely heuristically. 

\begin{figure}[!htb]
\begin{center}
	\includegraphics[width=9.4cm]{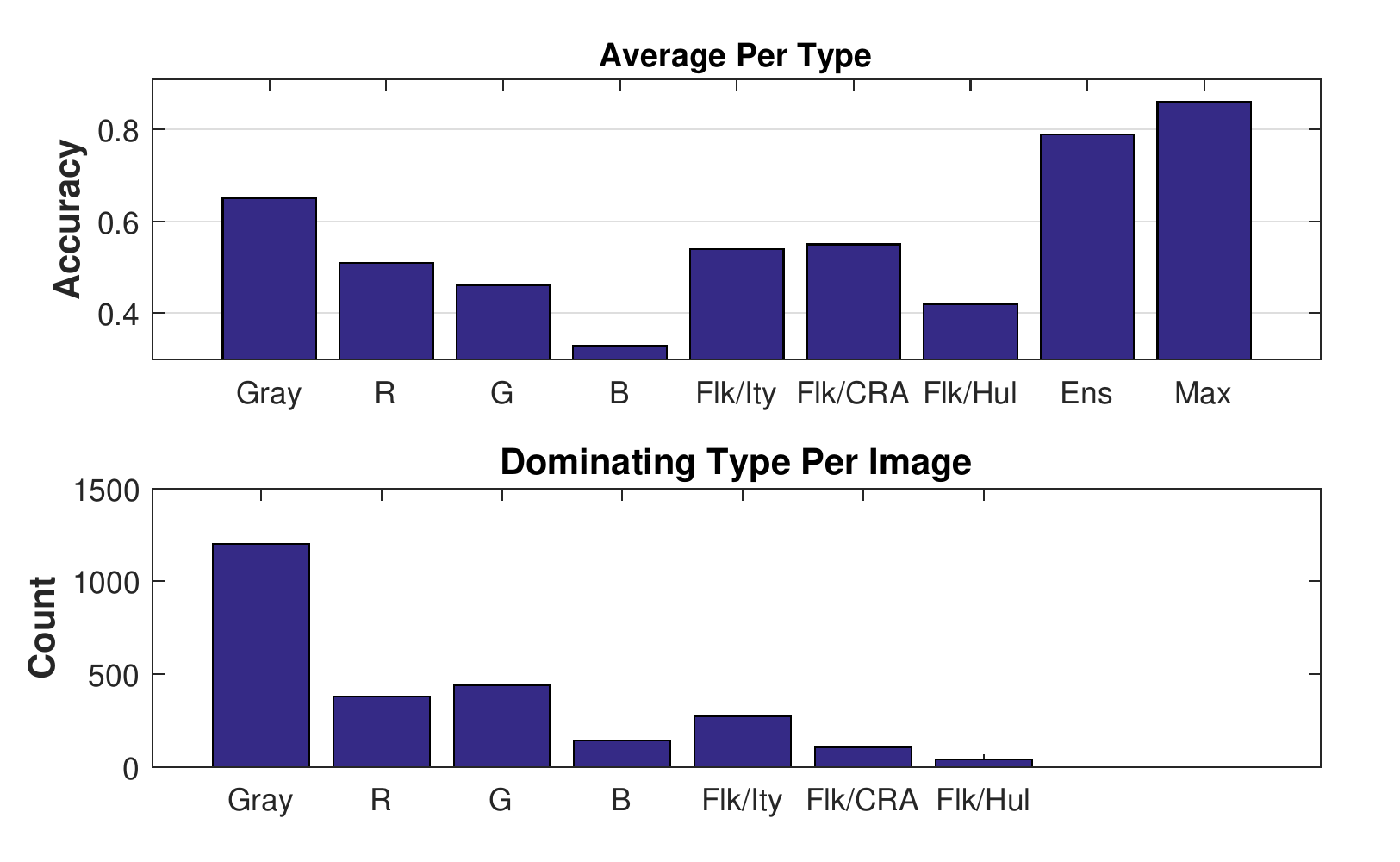}
	\caption{Accuracy of individual types (accuracy = sensitivity$\times$specificity). \textbf{Top Graph}: Average for all images for individual types (first 7 bars: gray-level, red, green, blue, interiority (\lesconfIty), based on \lesconfCRA, based on \lesconfHull; last two bars: ensemble and maximum per image. \textbf{Bottom Graph}: Count of dominating type per image (maximum accuracy for each type). }
	\label{fig_AccModes} 
\end{center}
\end{figure}

The (final) average accuracy per type is displayed in Fig. \ref{fig_AccModes}. Gray dominates in average, followed by the region types selected by \lesconfCRA~and \lesconfIty. The ensemble for all types (see label `Ens') - the thresholded, integrated map - is at an accuracy of 0.79, thus 0.12 better than its best performing type (gray). If we take the maximum type value per image, the accuracy arrives at 0.86 (right most bar, labeled `Max').

The bottom graph of Fig. \ref{fig_AccModes} shows a count of the dominating type per image. Gray now dominates even more, however the cluster types fall behing the chromatic types. 

% BF -----------------------------     ACCURACY MODES
% EF -----------

% SSSSSSSSSSSSSSSSSSSSSSSSSSSSSSSSSSSSSSSSSSSSSSSSSSSSSSSSSSSSSSSSSSSSSSSSSSSSSSSSSSSSSSSSSSSSSS
%										CLASSIFICATION										
% SSSSSSSSSSSSSSSSSSSSSSSSSSSSSSSSSSSSSSSSSSSSSSSSSSSSSSSSSSSSSSSSSSSSSSSSSSSSSSSSSSSSSSSSSSSSSS
\section{Classification}\label{s_Classification}

Descriptors from two sources are generated. One source are the regions as obtained with pixel-clustering used for segmentation (introduced above): those regions are described by their interior and their exterior (boundary) separately, each one by several attributes (Section \ref{ss_FeatClustRegion}). Another source is the topology analysis of the intensity image, which corresponds to the convential contour and region description in a scale space (Section \ref{ss_FeatTopo}). 

% SSSSSSSSSSSSSSSSSSSSSSSSSSSSS 	Features from Cluster Regions 	SSSSSSSSSSSSSSSSSSSSSSSSSSSSSSSSS
\subsection{Descriptors from Cluster Regions}\label{ss_FeatClustRegion}

Three types of descriptors are generated. One describes the region boundary. The other two describe the region interior, of which one regards the region pixels as a distribution and is therefore rather statistical in nature; the other describes the interior by a structural analysis with the symmetric-axis tansform.

\subsubsection{Boundary} % rad  cir  elo  cncv  Bis  star
For each region (obtained from \kmeans~clustering), its (closed) boundary is taken. The boundary is then transformed into a radial signature of which two groups of features are extracted. One group is derived from an analysis of the extrema in the signature, describing so the corners of the boundary. Another group are Fourier descriptors of the radial signature. Together ca. 8 parameters are generated.

% FLKSHAPE, f_FleckChar: 
% D.GEO   = [AreN Cmpc Lon Tri Qud   Star3 Star4];    % [nF 7] 	+ Appearances

\subsubsection{Pixel-Distribution}
Some regions of the lesion interior appear quasi-fragmented with large indentations and many holes, even when contiguous (see again Fig. \ref{fig_SegmDiff}). We assumed that those could be characteristics too for some cases. Attributes developed are the degree of solidity, compactness, `silhouetteness', centrality, peripherality, coverage, hollowness and `ringness'. 

%Att     = {'sold' 'cmpc' 'silh' 'cntr' 'peri' 'cvrg' 'holl' 'ring'};

\subsubsection{Structure of Interior}\label{sss_StructOfInterior}
The symmetric-axis transform is applied as described in \cite{Rasche2018_region}. The sym-axes are then partitioned at their branch-points and four types of descriptors are developed: short, long, forks and peaks. Short descriptors are a parameterization of short sym-axes segments and represent therefore bulky regions. Long descriptors represent long sym-axes segments and therefore represent elongated structures. Fork descriptors represent intersections of sym-axes segments. Peaks represent the immediate surround at local maxima in the distance map.

% SSSSSSSSSSSSSSSSSSSSSSSSSSSSS 	Features from Topology 	SSSSSSSSSSSSSSSSSSSSSSSSSSSSSSSSS
\subsection{Descriptors from Topology}\label{ss_FeatTopo}

Due to the relatively large image size, only two scales are generated, namely for $\sigma=1$ and $2$. Contours are extracted from both scales; regions only from the difference of the two.

\subsubsection{Contours}
Three types of contours are extracted, ridge, river and edge contours \cite{rasche2017rapid}. They are then partitioned and described using a local-to-global amplitude space \cite{Rasche10_covi}. Each partitioned segment is described by attributes such as geometric parameters, e.g. arc length, curvature, jaggedness, etc. as well as appearance parameters taken from the pixel values along the segment, e.g. mean intensity level, mean range value (contrast), the standard deviation ($\approx$ fuzziness), etc. Color information is included as well. In total, 15 attributes are used to describe a contour segment: 9 geometric parameters and 6 appearance parameters. Every contour type (ridge, river, edge) is described by the same set of parameters, arriving at 45 parameters for all contour types.

\subsubsection{Contour Groups}
Two types of contour groups are formed, clots and bundles. Clots are groups of short segments, whose endpoints lie near and therefore represent star-like groups. Bundles are groups of longer segments, whose midpoints lie near. 
The alignment analysis focuses in particular on radial and orientation statistics. For both clots and bundles, 12 geometric attributes and 7 appearance attributes are generated.

\subsubsection{Regions from Scale Space}
Regions are obtained from two sources: from the DOG map itself, as well as the region between edge contours from both scales. Region description is based on the interior as mentioned above (Section \ref{sss_StructOfInterior}).

%symmetric-axes transform \cite{Blum1973}, with an implementation as described in \cite{Rasche2018_region}.

For more details of this descriptor extraction process, we refer to \cite{Rasche2018satimg}.

% As the typical descriptor is expressed with around 10 or more attributes, the resulting histogram vector for a single descriptor contains 100 or more dimensions (110 for contour descriptors). 

%Thus, there is no use of the multi-dimensionality of the individual vectors per se; the histogram is a mere statistical description of the descriptor attributes present in an image and can be regarded as an analogue to 'image vectors', but not one for pixels yet one for attributes. It is the simplest type of representation one can think of when using lists of descriptors. 

% SSSSSSSSSSSSSSSSSSSSSSSSSSSSS 	RESULTS 	SSSSSSSSSSSSSSSSSSSSSSSSSSSSSSSSS
\subsection{Results}

The descriptor extraction output returns ca. 280 attributes in total, organized in lists of ca. 22 descriptor types. Approximately two third are geometric parameters, the remaining are appearance parameters. The simplest type of classification is histogramming of all attributes, thus ignoring the multi-dimensional representation of the descriptor; it can be taken as a lower performance bound. For a list of descriptors for an image, a 12-bin histogram $H_d$ is generated for each attribute $d$. Those attribute histograms (for all different descriptor types) are then concatenated to form a high-dimensional vector \hstAtt, whose size can more than three thousand dimensions (280 attributes $\times$ 12 bins). The combination of the principal component analysis (PCA) with a linear discriminant analysis (LDA) performed very well to separate those attribute histograms. 

\begin{figure}[!htb]
\begin{center}
	\includegraphics[width=7.6cm]{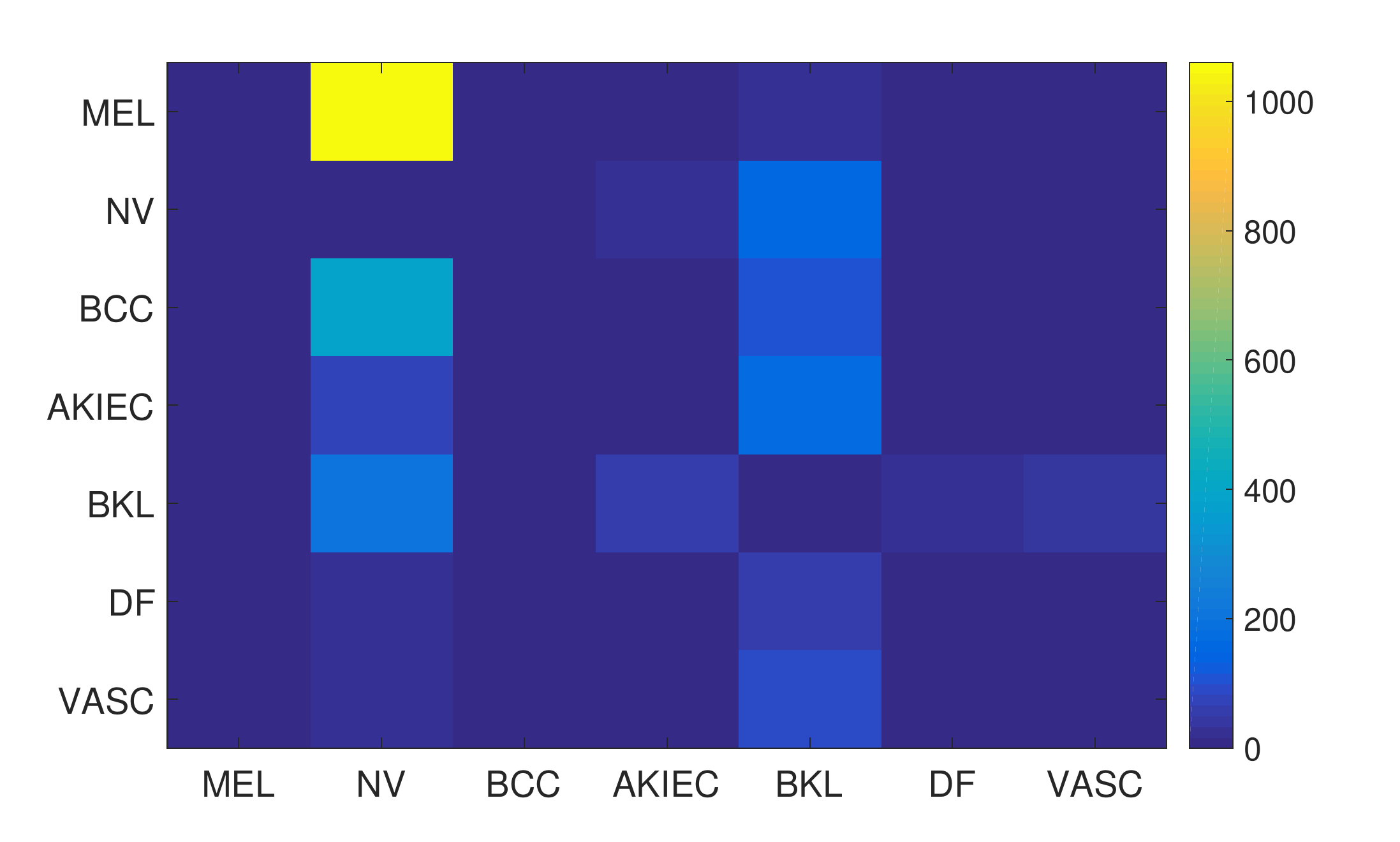}
	\caption{Confusion matrix for the 7 classes. The diagonal values are set to zero to highlight the confusions.}
	\label{fig_ConfMx} 
\end{center}
\end{figure}

On the training set, the histogramming approach achieved a prediction accuracy of 72.1 percent with a 5-fold cross-valdiation. Fig. \ref{fig_ConfMx} shows the confusion map. An analysis of the accuracies for individual descriptor types did produce very similar accuracies in the range between 67 and 69 percent. 

% BF -----------------------------     CONFUSION MATRIX
% EF -----------------------------

% SSSSSSSSSSSSSSSSSSSSSSSSSSSSSSSSSSSSSSSSSSSSSSSSSSSSSSSSSSSSSSSSSSSSSSSSSSSSSSSSSSSSSSSSSSSSSS
%										DISCUSSION										
% SSSSSSSSSSSSSSSSSSSSSSSSSSSSSSSSSSSSSSSSSSSSSSSSSSSSSSSSSSSSSSSSSSSSSSSSSSSSSSSSSSSSSSSSSSSSSS
\section{Discussion}\label{s_Discussion}

The segmentation results suggest (Fig. \ref{fig_AccModes}), that the gray-level is potentially one of the prevailing cues for doctors when observing the lesion boundaries; or perhaps the different analysis strategies bear the gray-scale level as the largest denominator.

We considered our method of thresholding the integration map only as a first step to combine the individiual segmentation types. In principal it has more potential, as the maximum value of the individual types per image is at an accuracy of 0.86 and thus 0.07 larger than the present strategy of thresholding the integrated maps (Fig. \ref{fig_AccModes}, upper graph). To our disappointment it was difficult to automatically predict the best performing types per image. It would however not be difficult to present to a doctor all different types and he would select which one appears to be the optimal one.

The segmentation approach involves minimal learning, namely storing only the RGB values for extracted regions that appear in the manually annotated regions. Apart from that, it is only a number of parameters that had to be adjusted. As such, the approach may therefore be suitable for fast localization of lesions.

That the classification results for individual descriptors show minor differences in accuracies hints that our approach has capture predominantly appearance aspects between classes and not structural differences. We still work on exploiting the multi-dimensional descriptor spaces.
%\section*{Acknowledgment}

%This work was fully supported by the Joint Applied Research Projects “Intelligent System for Automatic Assistance of Cervical Cancer Diagnosis”, grant number: PN-II-PT-PCCA-2013-4-0202, funded by Executive Unit for Higher Education, Research, Development and Innovation Funding (UEFISCDI).

\bibliographystyle{IEEEtran}
\bibliography{c:/klab/do/bib/MedSkin,c:/klab/do/bib/MedicalGeneral,c:/klab/do/bib/learn/networks,c:/klab/do/bib/covi,c:/klab/do/bib/rasche_ea,c:/klab/do/bib/books}

\begin{thebibliography}{10}
\providecommand{\url}[1]{#1}
\csname url@rmstyle\endcsname
\providecommand{\newblock}{\relax}
\providecommand{\bibinfo}[2]{#2}
\providecommand\BIBentrySTDinterwordspacing{\spaceskip=0pt\relax}
\providecommand\BIBentryALTinterwordstretchfactor{4}
\providecommand\BIBentryALTinterwordspacing{\spaceskip=\fontdimen2\font plus
\BIBentryALTinterwordstretchfactor\fontdimen3\font minus
  \fontdimen4\font\relax}
\providecommand\BIBforeignlanguage[2]{{%
\expandafter\ifx\csname l@#1\endcsname\relax
\typeout{** WARNING: IEEEtran.bst: No hyphenation pattern has been}%
\typeout{** loaded for the language `#1'. Using the pattern for}%
\typeout{** the default language instead.}%
\else
\language=\csname l@#1\endcsname
\fi
#2}}

\bibitem{codella2018ISIC}
N.~C. Codella, D.~Gutman, E.~Celebi, B.~Helba, M.~Marchetti, S.~W. Dusza,
  A.~Kalloo, K.~Liopyris, H.~Kittler, N.~Mishra, and A.~Halpern, ``Skin lesion
  analysis toward melanoma detection: A challenge at the 2017 international
  symposium on biomedical imaging ({ISBI}), hosted by the international skin
  imaging collaboration ({ISIC}),'' \emph{arXiv preprint arXiv:1710.05006},
  2017.

\bibitem{tschandel2018ham}
P.~Tschandl, C.~Rosendahl, and H.~Kittler, ``The {HAM}10000 dataset: A large
  collection of multi-source dermatoscopic images of common pigmented skin
  lesions,'' \emph{arXiv:1803.10417}, 2018.

\bibitem{korotkov2012computerized}
K.~Korotkov and R.~Garcia, ``Computerized analysis of pigmented skin lesions: a
  review,'' \emph{Artificial intelligence in medicine}, vol.~56, no.~2, pp.
  69--90, 2012.

\bibitem{scharcanski2014computer}
J.~Scharcanski and M.~E. Celebi, \emph{Computer vision techniques for the
  diagnosis of skin cancer}.\hskip 1em plus 0.5em minus 0.4em\relax Springer,
  2014.

\bibitem{mishra2016overview}
N.~K. Mishra and M.~E. Celebi, ``An overview of melanoma detection in
  dermoscopy images using image processing and machine learning,'' \emph{arXiv
  preprint arXiv:1601.07843}, 2016.

\bibitem{celebi2013lesion}
M.~E. Celebi, Q.~Wen, S.~Hwang, H.~Iyatomi, and G.~Schaefer, ``Lesion border
  detection in dermoscopy images using ensembles of thresholding methods,''
  \emph{Skin Research and Technology}, vol.~19, no.~1, pp. e252--e258, 2013.

\bibitem{neghinua2016automatic}
C.~Neghin{\u{a}}, M.~Zamfir, A.~Sultana, E.~Ovreiu, and M.~Ciuc, ``Automatic
  detection of hemangiomas using unsupervised segmentation of regions of
  interest,'' in \emph{Communications (COMM), 2016 International Conference
  on}.\hskip 1em plus 0.5em minus 0.4em\relax IEEE, 2016, pp. 69--72.

\bibitem{barata2014two}
C.~Barata, M.~Ruela, M.~Francisco, T.~Mendon{\c{c}}a, and J.~S. Marques, ``Two
  systems for the detection of melanomas in dermoscopy images using texture and
  color features,'' \emph{IEEE Systems Journal}, vol.~8, no.~3, pp. 965--979,
  2014.

\bibitem{Rasche2018satimg}
\BIBentryALTinterwordspacing
C.~Rasche, ``Deep structural analysis for high-resolution images,'' \emph{IEEE
  Geo Sci.}, vol. under review, 2018. [Online]. Available:
  \url{https://www.researchgate.net/publication/323756034_Deep_Structural_Analysis_for_High-Resolution_Images}
\BIBentrySTDinterwordspacing

\bibitem{Rasche2018_region}
\BIBentryALTinterwordspacing
------, ``Rapid region analysis for classification,'' vol. under review.
  [Online]. Available:
  \url{https://www.researchgate.net/publication/322570930_Rapid_Region_Analysis_for_Classification}
\BIBentrySTDinterwordspacing

\bibitem{rasche2017rapid}
------, ``Rapid contour detection for image classification,'' \emph{IET Image
  Processing}, vol.~12, no.~4, pp. 532--538, 2017.

\bibitem{Rasche10_covi}
------, ``An approach to the parameterization of structure for fast
  categorization,'' \emph{International Journal of Computer Vision}, vol.~87,
  pp. 337--356, 2010.

\end{thebibliography}

\end{document}